# Optimizing Predictive AI in Physical Design Flows with Mini Pixel Batch Gradient Descent

Haoyu Yang  
NVIDIA Corp.

Anthony Agnesina  
NVIDIA Corp.

Haoxing Ren  
NVIDIA Corp.

**Abstract**

Exploding predictive AI has enabled fast yet effective evaluation and decision-making in modern chip physical design flows. State-of-the-art frameworks typically include the objective of minimizing the mean square error (MSE) between the prediction and the ground truth. We argue the *averaging* effect of MSE induces limitations in both model training and deployment, and good MSE behavior does not guarantee the capability of these models to assist physical design flows which are likely sabotaged due to a small portion of prediction error. To address this, we propose mini-pixel batch gradient descent (MPGD), a plug-and-play optimization algorithm that takes the most informative entries into consideration, offering probably faster and better convergence. Experiments on representative benchmark suits show the significant benefits of MPGD on various physical design prediction tasks using CNN or Graph-based models.

## 1 Introduction

Exploding predictive AI techniques have enabled fast yet effective evaluation and decision-making in back-end chip design flows. The practical methodology is training predictive machine learning models as replacements for costly heuristic algorithms used in physical design feedback loops [4, 6, 8, 10, 14, 16, 23, 27]. For example: In routability-driven placement, a congestion or overflow map generated by a machine-learning model will be used to provide additional constraints during the placement runtime [7]; In mask optimization, an AI-based lithography simulator performs fast aerial image and contour estimation that will guide design calibration [5, 13]; A fast timing prediction engine can also enable shorter turn-around-time during design optimization [10]. The performance of predictive models therefore has a significant impact on the final QoR of a chip design. During the placement runtime, cells are usually inflated based on the congestion estimation and hence cells in high congested regions can be further spread away. However, minor prediction errors in the congestion map will even result in 65% more routing overflow[1]. In terms of mask optimization, 1% intensity error in a single pixel of the aerial image can lead to >5$nm$ final contour edge displacement and hence cause yield loss [5].

Many attempts are made to further improve the performance of predictive AI. Major methodologies include: improving the data quality and quantity, reforming the model architectures that are correctly biased with domain knowledge, and developing better training algorithms. However, we focus on the training algorithm and propose general solutions for the predictive AI in physical design flows. State-of-the-art predictive AI training solutions fall into a common objective, that is, minimizing the MSE between the prediction and the ground truth. We argue that the native MSE cost function might not be the best choice under large-scale regression application scenarios for the following reasons: **First, MSE cannot accurately monitor the model performance.** MSE computes the average error across the entire output entries, which could be extremely small at a later training state even if there exists an "outlier" pixel that differs from the ground truth by a large amount. **Second, MSE may cause the model to stuck at local minima under highly non-convex circumstances.** The *averaging* effect of MSE makes the gradient small in each updating step hence preventing the optimizer from exploring a wider solution space.

---
[1]Tested on superblue11 [2] using DREAMPlace [16].

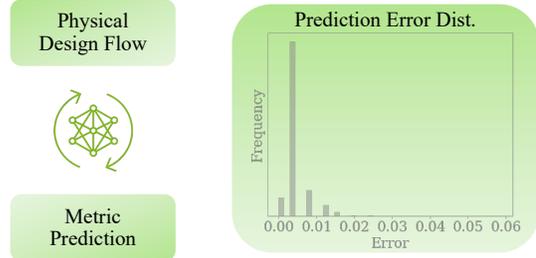

Figure 1: Predictive AI plays significant roles in the chip design flow. Example experiments on lithography modeling tasks show models trained with MSE achieve accurate prediction on over 95% of the data entries, leaving a small fraction yet important entries unattended with fetal error.

Considering the potential limitations of MSE come from treating the prediction error of each data entry evenly, an appropriate cost function should pay more attention to the most informative data points and entries in each data point, which can provide effective gradients for modeling training. However, most research focuses on the former term and there are limited investigations on the in-data entries. Recent research presents ideas that place a modulator on the original loss function and allow the optimizer to focus more on the samples with large prediction errors. DLHSD [24] is one of the earliest attempts using biased loss to penalize easy-to-classify hotspot instances. Shrinkage loss [18] re-weights the squared loss $L_2$ such that the optimizer can focus on hard samples in image translation tasks. Very recently, CircuitNet [4] proposed a biased loss that has a similar structure as shrinkage loss. The only difference is that [4] penalizes the $L_2$ loss with respect to the individual pixel value of the ground truth image. However, these representative solutions are still a soft penalty on the cost function, which fluctuates along with the training procedure and is not as stable as traditional MSE. To address this problem, we present mini pixel batch gradient descent (MPGD), an optimization flow that incorporates an adaptive mean square error and active sampling. The solution takes most informative data entries into consideration and places constraints to focus on the largest prediction error by ignoring partial of the entries in the training target, while maintaining a stable and faster convergence. Our major contributions include:

- We discuss the recent development of predictive AI in chip design flows and the limitations of commonly used MSE cost function, inspired by which, an adaptive mean square error (AMSE) cost function is proposed.
- We propose in-data active sampling (IAS) of critical entries to take full advantage of AMSE for fast and stable convergence.
- We introduce MPGD, a plug-and-play training algorithm that honors prediction error of critical entries through AMSE and IAS. We also show provably faster convergence of MPGD than traditional solutions.
- We conduct experiments on lithography, timing, and routability prediction tasks and demonstrate that MPGD has general improvements in predictive AI models (CNN, Physics-Inspired NN, and GCN) for back-end chip design tasks.

The rest of the manuscript is organized as follows: Section 2 introduces related works of predictive AI in back-end chip design flow and task-specific optimization solutions; Section 3 covers the details of the development of MPGD and supporting analysis; Section 4 show the effectiveness of the proposed MPGD algorithm on recent open chip design benchmarks followed by the conclusion in Section 5.

## 2 Related Works
### 2.1 Predictive AI In Chip Design Flow

Major application scenarios of predictive AI in chip design flow fall into performance evaluation and decision-making. Representative works include routability estimation [4, 17, 23] for chip placement, lithography modeling [6, 14, 26, 28] for reticle enhancement, sign-off timing prediction [9, 10, 22] for routing optimization, and so on. The machine learning backbones include CNN (UNet [21], ResNet [12]), Graph (GCN[15], GraphSAGE[11]) and multimodals [22]. These frameworks usually take intermediate data packets from the physical implementation flow as inputs and perform predictions on PPA at the output, which will be used to provide a feedback loop for design optimization. In this paper, we will focus on improving the performance of both CNN and Graph-based regression models in chip design flow.

### 2.2 Reweighted MSE

Several attempts are mitigating the drawbacks of MSE. Shrinkage loss [18] is one of the effective approaches. The beneath idea of shrinkage loss is loss reweighting and amplifying the instances with higher prediction error, which is similar to employing higher order norms as cost functions. Shrinkage loss can be expressed as follows:

$$l_s = \frac{l^2}{1 + \exp(a \cdot (c - l))}, \quad (1)$$

where $l$ is the original $L_2$ loss, $a$ and $c$ are some user-defined coefficients. Note that shrinkage loss applies to both training instances or entries of one instance. Very recently, a biased loss [4] is also proposed for MSE reweighting purposes. Unlike shrinkage loss, biased loss reweights the loss according to the magnitude of the ground truth value (Equation (2)).

$$l_{\text{bias}} = \frac{l^2}{1 + \exp(a \cdot (c - \max \boldsymbol{y}^*))}. \quad (2)$$

There are also alternatives by capturing the top-k instances with the largest error through a differentiable way [20]. However, existing solutions are mostly applying soft penalties on the original MSE which did not resolve the problem efficiently, and our MPGD outperforms these approaches with provably better convergence.

## 3 The Algorithm
### 3.1 Adaptive Mean Square Error

In typical predictive AI problems, MSE and its variations are well-accepted cost functions that are used to minimize the difference between the prediction and the ground truth target. One of the virtues of MSE is its capability to focus on worst-case scenarios across data points. However, this is not always true and there are some drawbacks of MSE when dealing with prediction tasks that are critical in down-stream flows.

Let $f : \mathbb{R}^m \to \mathbb{R}^n$ be some predictive model parameterized by $\boldsymbol{w}$ with $\boldsymbol{y} = f(\boldsymbol{x}; \boldsymbol{w})$, and $\boldsymbol{y}^*$ is the corresponding ground truth. The MSE-based training cost function is usually computed as

$$l = \frac{1}{n} \sum_{i=1}^{n} (y_i - y_i^*)^2, \quad (3)$$

where $i$ indicates some entry of the output vector.

First, we argue that Equation (3) is not a good indicator of the quality of the predictive model under certain scenarios, especially when $\boldsymbol{y}$ is identical to $\boldsymbol{y}^*$ at almost every entry other than few locations $\mathcal{E} = \{e_1, e_2, ..., e_k\}$, where $y_{e_i}$ differs from $y_{e_i}^*$ by some reasonable amount. Here we define $\mathcal{E}$ as the set of critical entries. Under this assumption, Equation (3) becomes,

$$l \approx \frac{1}{n} \sum_{e_i \in \mathcal{E}} (y_{e_i} - y_{e_i}^*)^2, \quad (4)$$

and $l$ is extremely small given $k = |\mathcal{E}| \ll n$, which indicates that the model is flawed even with good MSE behavior. *This usually happens at a later training stage when the training loss is small and the optimizer makes little progress in each epoch.* Similarly, from an optimization point of view, it is hard to further push down the errors in the scenario of Equation (4). For simplicity, we use naive gradient descent as an example, where

$$\frac{\partial l}{\partial \boldsymbol{w}} = \frac{2}{n} \sum_{e_i \in \mathcal{E}} (y_{e_i} - y_{e_i}^*) \cdot \frac{\partial y_{e_i}}{\partial \boldsymbol{w}}, \quad (5)$$

which has a limited impact on the model training given $k \ll n$.

An intuitive workaround for these concerns is amplifying these minority error entries as much as possible, that is

$$l = ||\boldsymbol{y} - \boldsymbol{y}^*||_\infty = \max |\boldsymbol{y} - \boldsymbol{y}^*|. \quad (6)$$

However, a machine learning model is hardly trainable through Equation (6), because only one entry of a data point is considered when computing the loss, resulting in unstable optimization. Though we can seek a compromise through higher order $L_p$ cost functions, we still either lose accurate measurement or stable convergence or both.

However, this is not a showstopper of finding better training strategies. There are two observations from the above discussion: (1) The drawbacks of MSE manifest at a later training stage and (2) The side impacts get worse when the prediction dimension $n$ is large. This inspires us to design the cost function called adaptive mean square error (AMSE):

$$l = \frac{1}{k} \sum_{e_i \in \mathcal{E}} (y_{e_i} - y^*_{e_i})^2, \quad (7)$$

where $k = |\mathcal{E}|$ is the number of entries that have a large prediction error. There are several virtues of AMSE:

- By the definition of $\mathcal{E}$, $l$ will only consider the entries that contribute most to the prediction error, making it an accurate estimator of the model performance.
- As long as there are critical errors in the prediction, $l$ will be large enough to generate gradient information during the training procedure.

It should also be noted that Equation (7) resembles the top-k error well used in ML community. However, AMSE applies entry-wise instead of instance-wise and $k$ **adaptively changes along with training**.

### 3.2 In-Data Active Sampling of Critical Entries

Equation (7) indicates that during training, we should focus on critical entries in each data point, and the choice of $\mathcal{E}$ has significant impact on the convergence. Essentially, we want the chosen $\mathcal{E}$ to be informative while maintaining a stable training procedure. Therefore, we follow the concept of expected model change [3] that is widely adopted in active learning flows and formulate the problem of finding proper $\mathcal{E}$ as an active learning procedure:

**Problem 1** (In-Data Active Sampling).

$$\max_{\mathcal{E}_t} |\mathcal{E}_t|, \quad (8)$$

$$\text{s.t. } ||\nabla l_t||^2 > \sigma_t, \quad (9)$$

where $t$ denotes the current training step, $\sigma_t$ is some predefined threshold term that controls the expected model change and $l_t$ represents the loss computed at time step $t$ using Equation (7).

Solving Problem 1 will grant that (a) $\mathcal{E}_t$ is informative to generate enough gradients for entries with large prediction error (by Equation (9)) and (b) the gradient will not be too large and causing training fluctuation (by Equation (8)). The reason for (a) is straightforward because we ask the gradient to be greater than some value. The intuition for (b) comes from the fact that a larger batch size usually yields slower and more stable gradient descent during neural network training. Since Problem 1 is solved in every optimization step, we need an efficient solution to avoid significant training overhead. Here we propose a heuristic that shares most of the original computations in the training flow (Algorithm 1): we first perform the forward computation to obtain the prediction and get the corresponding absolute error of each entry (lines 1–2), and then we find the indices where the absolute error is greater than some predefined threshold (line 3).

It should be noted that Algorithm 1 does not solve Problem 1 and honor constraints directly. This is because evaluating Equation (9) and determining proper $\sigma_t$ are costly. However, we assume that $\eta_t$

**Algorithm 1** In-Data Active Sampling

**Input:**  Model $f$, training data pair $\{x, y^*\}$, sampling constant $\lambda$;
**Output:** Critical entry indices $\mathcal{E}_t$;
1: Forward computing $y \leftarrow f(x)$;
2: $d_{\text{AE}} \leftarrow \dfrac{|f(x) - y^*|}{\max y^*}$;
3: $\mathcal{E}_t \leftarrow$ Get indices that satisfy $d_{\text{AE}} > \lambda$;
4: **if** $\mathcal{E}_t = \emptyset$ **then**
5: $\quad \mathcal{E}_t \leftarrow$ All indices of the training data;

**Algorithm 2** Mini Pixel Batch Gradient Descent

**Input:** $f$ parameterized by $w$, training data set $\{\mathcal{X}, \mathcal{Y}\}$, critical threshold $\lambda$;
**Output:** Trained model $w$;
1: **for** Total Number of Training Steps **do**
2: $\quad$ Sample a batch of $n$ training data pairs $\{x_i, y^*_i, i = 1, 2, ..., n\}$;
3: $\quad$ **for** Each instance $i$ in the batch **do**
4: $\quad\quad$ Get critical entry indices $\mathcal{E}_i$ by Algorithm 1;
5: $\quad\quad$ $l_i \leftarrow$ Compute AMSE by Equation (7);
6: $\quad$ $l \leftarrow \frac{1}{n} \sum_{i=1}^{n} l_i$;
7: $\quad$ $w \leftarrow$ Backward step and update parameters.

reduces along with model training and converging to ensure there exists a $\mathcal{E}_t$ that satisfies Equation (9), which is also guaranteed by line 3. In addition, we can avoid computing traditional MSE loss by introducing a predefined constant $\lambda$.

### 3.3 Mini Pixel Batch Gradient Descent

The AMSE and IAS comprise the key ideas of the mini-pixel batch gradient descent. We summarize the details of MPGD in Algorithm 2: Following regular neural network training, MPGD works in a batch manner (line 1); In each training step, we sample a batch of training instances and compute their respective critical entries and AMSE (lines 2–5); Finally, neuron weights are updated by passing the gradient of AMSE backward (lines 6–7).

*3.3.1 Advantages of MPGD.* Compared to traditional MSE-based optimization, MPGD has several benefits. First, MPGD enjoys stable convergence similar to MSE-based solutions. Because the model behaves poorly at early training stages, and $\mathcal{E}$ covers all the entries of the training data, which makes Equation (7) the same as traditional MSE. Second, as the model converges, the optimizer will pay more attention to the entries with large prediction errors, circumventing the "averaging" effect of the traditional MSE-based approach. Finally, in most optimization steps, MPGD will only consider a fraction of the entries of each data, posing a pseudo-stochastic effect that helps the optimizer jump out of the local minimum. This is especially effective when the model is highly non-convex.

*3.3.2 Convergence of MPGD.* By intuition, MPGD is expected to perform better than MSE-based optimization through faster and better convergence, which is also quantitatively true. For a convex problem, convergence usually means global optimum when $\nabla l = 0$, where $\nabla l = \dfrac{1}{n} \sum_{i=0}^{n-1} \nabla l_i$, $l_i$ is the loss of each entry in the data and $N$ is the total number of entries. $l$ denotes some cost function. This

Conference'17, July 2017, Washington, DC, USA       Haoyu Yang, Anthony Agnesina, and Haoxing Renis however not achievable under most neural network scenarios due to non-convexity. Here we relax the convergence condition to

$$||\nabla l||^2 < \epsilon, \tag{10}$$

for some super small $\epsilon$. Due to the same reason that MSE cannot accurately monitor the model performance, we focus on the top-k critical entries

$$||\nabla l_{\text{topK}}||^2 = ||\frac{1}{k}\sum_{i \in \text{topK}} \nabla l_i||^2 < \epsilon, k \ll n, \tag{11}$$

where we denote $l_{\text{topK}}$ as the loss computed by Equation (7). Under the premise $k \ll n$ and Equation (5), it is safe to derive that $||\nabla l_{\text{topK}}||^2 \geq ||\nabla l||^2$. Let

$$||\nabla l_{\text{topK}}||^2 = \eta ||\nabla l||^2, \eta \geq 1. \tag{12}$$

We will show the convergence rate of MPGD versus gradient descent using the original MSE in the following deduction.

**Lemma 1** (Descent Lemma [19]). *Assume $\nabla l$ is Lipschitz continuous with constant L, each step of MSE-based gradient descent is bounded as follows:*

$$l_t \leq l_{t-1} - \frac{1}{2L}||\nabla l_{t-1}||^2, \tag{13}$$

*where t indicates the optimization step.*

**Theorem 1.** *Assume $\nabla l$ is Lipschitz continuous with constant L and $\nabla l$ is bounded, MSE-based gradient descent achieves Equation (11) at the rate of $\mathcal{O}(\frac{\eta(l_0-l^*)}{\epsilon})$, where $l_0$ is the initial loss and $l^*$ corresponds to the lower bound of the loss function computed through Equation (3).*

Proof. According to Lemma 1,

$$||\nabla l_{t-1}||^2 \leq 2L(l_{t-1} - l_t). \tag{14}$$

Taking the summation on both side yields,

$$\sum_{t=1}^{T} ||\nabla l_{t-1}||^2 \leq 2L(l_0 - l_1 + l_1 - l_2 + ... + l_{T-1} - l_T)$$
$$= 2L(l_0 - l_T) \leq 2L(l_0 - l^*). \tag{15}$$

Obviously,

$$\sum_{t=1}^{T} ||\nabla l_{t-1}||^2 \geq T \min_{t \in [0,T-1]} \{||\nabla l_t||^2\}. \tag{16}$$

Substitute Equation (16) into Equation (15), we have

$$T \min_{t \in [0,T-1]} \{||\nabla l_t||^2\} \leq 2L(l_0 - l^*), \tag{17}$$

i.e.

$$\min ||\nabla l||^2 = \min \frac{||\nabla l_{\text{topK}}||^2}{\eta} \leq \frac{2L(l_0 - l^*)}{T}, \tag{18}$$

and hence

$$\min ||\nabla l_{\text{topK}}||^2 \leq \frac{2\eta L(l_0 - l^*)}{T}. \tag{19}$$

Equation (19) indicates that at most $T$ optimization steps, there exists one timestamp that the gradient norm is upper bounded by $\frac{2\eta L(l_0-l^*)}{T}$. Let

$$\epsilon \geq \frac{2\eta L(l_0 - l^*)}{T} \Rightarrow T \geq \frac{2\eta L(l_0 - l^*)}{\epsilon}. \tag{20}$$

**Table 1: Benchmarks used throughout the experiments.**

| Benchmarks | Task | Technology | Backbone |
|---|---|---|---|
| LithoBench [28] | Lithography | NanGate45 | FNO, CNN |
| CircuitNet 2.0 [4] | Congestion/DRV | N28 | CNN |
| OpenROAD [1] | Timing | SkyWater130 | GCN |

**Table 2: Experimental results on aerial image prediction.**

| Method | Metal | | | Via | | |
|---|---|---|---|---|---|---|
| | SSIM↑ | MSE↓ | ME↓ | SSIM↑ | MSE↓ | ME↓ |
| DOINN [25] | 0.995 | 0.002 | 0.066 | 0.673 | 0.025 | 0.975 |
| w/ Shrinkage | 0.991 | 0.002 | 0.138 | 0.999 | 0.001 | 0.039 |
| w/ MPGD | 0.990 | 0.002 | **0.042** | 0.997 | 0.001 | **0.035** |
| DAMO [5] | 0.998 | 0.001 | 0.018 | 0.999 | 0.000 | 0.015 |
| w/ Shrinkage | 0.996 | 0.001 | 0.020 | 0.999 | 0.001 | 0.021 |
| w/ MPGD | 0.997 | 0.001 | **0.017** | 0.999 | 0.000 | **0.013** |

**Table 3: Experimental results on arrival time prediction performance ($R^2 \uparrow$), compared to OpenROAD [1] STA.**

| Design | DAC'22 [10] | w/Shrinkage | w/MPGD |
|---|---|---|---|
| jpeg_encoder | 0.882 | 0.934 | **0.987** |
| usbf_device | 0.925 | 0.919 | **0.944** |
| aes192 | 0.861 | **0.939** | 0.931 |
| xtea | 0.913 | 0.956 | **0.984** |
| spm | 0.897 | 0.944 | **0.983** |
| y_huff | 0.926 | 0.911 | **0.966** |
| synth_ram | 0.866 | 0.751 | **0.935** |
| Average | 0.896 | 0.907 | **0.961** |

This implies that we will need $\mathcal{O}(\frac{\eta(l_0-l^*)}{\epsilon})$ steps to satisfy the convergence condition in Equation (11). □

Similarly, we can derive the convergence rate of MPGD.

**Theorem 2.** *Assume $\nabla l$ is Lipschitz continuous with constant L and $\nabla l$ is bounded, MPGD achieves Equation (11) at the rate of $\mathcal{O}(\frac{\eta(l_0-l^*_{topK})}{\epsilon})$, where $l_0$ is the initial loss and $l^*_{topK}$ corresponds to the lower bound of the loss function computed through Equation (7).*

Proof. Replacing the $l$'s with their corresponding $l_{\text{topK}}$ in the proof of theorem 1 gives

$$T \geq \frac{2\eta L(l_{0,\text{topK}} - l^*_{\text{topK}})}{\epsilon} \approx \frac{2\eta L(l_0 - l^*_{\text{topK}})}{\epsilon}, \tag{21}$$

assuming the model behaves poorly on all the entries at the beginning such that $l_0 \approx l_{0,\text{topK}}$. □

*3.3.3 Choice of $\lambda$.* The deduction above quantitatively shows that MPGD converges $(\frac{l_0-l^*}{l_0-l^*_{\text{topK}}})\times$ faster than traditional MSE-based gradient descent. We can also notice that $l^*_{\text{topK}}$ is mainly controlled by $\lambda$ in Algorithm 1 and a larger $\lambda$ implies larger benefits of MPGD. However, there is a strong positive correlation between $\lambda$ and $\eta$. Thus, we will not benefit too much from increasing $\lambda$, because this will in general slow down the convergence. Empirically, we choose $\lambda$ between $10^{-2} - 10^{-3}$.

## 4 Experiments

### 4.1 Benchmarks

In this section, we will demonstrate the effectiveness of MPGD by plugging it into four predictive AI frameworks. All benchmarks are obtained from open-source PDK and implementation flows, which are summarized in Table 1. In particular, we use the LithoBench [28] dataset from lithography aerial image simulation, CircuitNet 2.0 [4] for routing congestion and design rule violation prediction and OpenROAD [1] test designs for timing prediction. The benchmark also covers a wide range of technology nodes and different AI backbones including CNN and GCN. *All experiments are conducted on the same platform using NVIDIA A100, where each model is trained from scratch using the same configurations as in their original paper except for the loss function. We also use a constant $\lambda = 0.007$ throughout the training.*

There are multiple metrics used to evaluate the performance of predictive AI in these tasks, which are defined as follows:

**Definition 1** (SSIM). The structural similarity index measure (SSIM) measures the similarity between two images.

$$\text{SSIM}(x, y) = \frac{(2\mu_x \mu_y + c_1)(\sigma_{xy} + c_2)}{(\mu_x^2 + \mu_y^2 + c_1)(\sigma_x^2 + \sigma_y^2 + c_2)}, \quad (22)$$

where $\mu$ is the pixel mean of an image, $\sigma^2$ denotes variance of an image, $\sigma_{xy}$ is the covariance and $c_1, c_2$ are coefficients that stablize the denominator.

**Definition 2** (MSE, NRMSE, peakNRMSE). There are multiple MSE-related metrics used in our experiments, with MSE being the traditional mean square error, and the others are defined on the top of MSE,

$$\text{NRMSE}(x, y) = \frac{1}{\max y - \min y} \sqrt{\text{MSE}(x, y)}, \quad (23)$$

and the peakNRMSE is given by the NRMSE between the top-k entries of $x$ and $y$.

**Definition 3** (ME). Max Error (ME) is the measurement of the worst prediction error in an image,

$$\text{ME}(x, y) = \frac{\max |x - y|}{\max y}. \quad (24)$$

**Definition 4** ($R^2$). The coefficient of determination ($R^2$) measures the performance of regression tasks which is defined as

$$R^2(x, y) = 1 - \frac{\sum_i (x_i - y_i)^2}{\sum_i (y_i - \bar{y})^2}. \quad (25)$$

### 4.2 Evaluation of MPGD Various Tasks

*4.2.1 Evaluation of Lithography Modeling* In this first experiment, we focus on the optical modeling of lithography simulation, that is predicting the aerial image intensity given a mask design. Here we adopt two popular architectures for lithography: DOINN [25] and DAMO [5]. The former is an optics-inspired model with the Fourier neural operator as the backbone and the latter is a standard CNN-based model. Three cost functions are examined on these architectures that include traditional MSE, shrinkage loss [18], and our proposed MPGD (Table 2). On both metal and via dataset, we show that even two candidate network architectures achieve superior results on aerial image prediction in terms of MSE, there is at least one pixel that has a very high aerial intensity error, which will induce significant resist contour error. While the proposed MPGD can achieve the best worst-case error thanks to the advantages of AMSE and ISA.

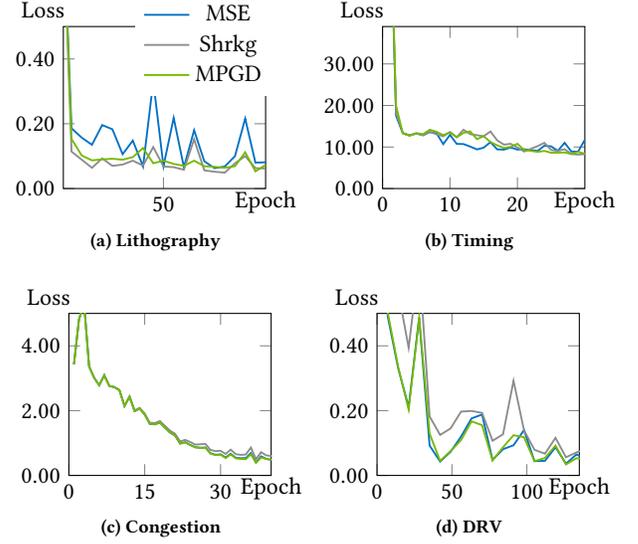

Figure 2: Training Visualization of different prediction tasks.

*4.2.2 Evaluation of Endpoint Arrival Time Prediction* Pre-route timing prediction is a critical step in the back-end design flow, which helps to circumvent the computing overhead of routing and static timing analysis, as in Table 3. We choose [10] as the baseline model for pre-route timing prediction, which uses graph convolution networks as its backbone. Similarly, the baseline experiment uses traditional MSE and we implement shrinkage loss and MPGD on the top of the baseline model. We use the same training and testing set as in [10], and we show that with the help of MPGD, there is a $\sim 9\%$ improvement in endpoint arrival time prediction.

*4.2.3 Evaluation of Routing Congestion Prediction* Routing congestion is one of the earliest AI applications in chip design flows. Originally, global routing steps were required to obtain accurate congestion information, which is costly because congestion estimation needs to be triggered frequently during the placement runtime. RouteNet [23] is the very first framework to perform routing congestion estimation with generative AI, which is backboned with fully convolutional networks (FCN). RouteNet with traditional MSE loss will be the baseline of our experiments. We also implemented the biased loss [4] and shrinkage loss as additional comparisons, as in Table 4. We can observe that biased loss exhibits an advantage in certain peakNRMSE settings over the traditional MSE loss function, it however induces degradation in other terms like SSIM. On the other hand, MPGD is able to achieve the best (or almost the best) results on all evaluation metrics achievable by different configurations of biased loss, which demonstrates the effectiveness of MPGD.

Table 4: Experimental results on routability prediction.

| Task | Setting | NRMSE ↓ | SSIM ↑ | peak NRMSE ↓ | | | | |
|---|---|---|---|---|---|---|---|---|
| | | | | 0.5% | 1% | 2% | 5% | average |
| Congestion | RouteNet [23] | 0.047 | 0.773 | 0.441 | 0.323 | 0.236 | 0.155 | 0.289 |
| | w/ biased loss ($s=10, b=0.4$) | 0.059 | 0.735 | 0.337 | 0.244 | 0.179 | **0.124** | 0.221 |
| | w/ biased loss ($s=20, b=0.4$) | 0.105 | 0.616 | 0.267 | **0.205** | 0.171 | 0.157 | 0.200 |
| | w/ biased loss ($s=30, b=0.4$) | 0.183 | 0.421 | **0.263** | 0.229 | 0.222 | 0.237 | 0.238 |
| | w/ biased loss ($s=20, b=0.2$) | 0.057 | 0.740 | 0.367 | 0.264 | 0.190 | 0.126 | 0.237 |
| | w/ biased loss ($s=20, b=0.6$) | 0.238 | 0.330 | 0.311 | 0.320 | 0.335 | 0.350 | 0.329 |
| | w/ Shrinkage | 0.050 | 0.758 | 0.274 | 0.220 | 0.173 | 0.123 | 0.197 |
| | w/ MPGD | **0.046** | **0.787** | 0.271 | 0.217 | **0.171** | 0.121 | **0.195** |
| DRV | RouteNet [23] | 0.014 | 0.742 | 0.068 | 0.056 | 0.046 | 0.036 | 0.052 |
| | w/ biased loss ($s=10, b=0.4$) | 0.025 | 0.533 | 0.095 | 0.084 | 0.072 | 0.058 | 0.077 |
| | w/ biased loss ($s=20, b=0.4$) | 0.228 | 0.021 | 0.382 | 0.382 | 0.371 | 0.317 | 0.363 |
| | w/ biased loss ($s=30, b=0.4$) | 0.554 | 0.004 | 0.490 | 0.499 | 0.508 | 0.523 | 0.505 |
| | w/ biased loss ($s=20, b=0.2$) | 0.031 | 0.418 | 0.099 | 0.089 | 0.079 | 0.063 | 0.082 |
| | w/ biased loss ($s=20, b=0.6$) | 0.624 | 0.003 | 0.609 | 0.620 | 0.626 | 0.625 | 0.620 |
| | w/ Shrinkage | 0.022 | 0.364 | 0.069 | 0.059 | 0.051 | 0.042 | 0.055 |
| | w/ MPGD | **0.012** | **0.772** | **0.068** | **0.056** | **0.045** | **0.034** | **0.051** |

*4.2.4 Evaluation of DRV Prediction* The last experiment is the prediction of design rule violations (Table 4). This is similar to the congestion prediction task except the model will take additional congestion information as inputs. We use the same RouteNet [23] framework as in previous experiments and the network architecture is adjusted to accommodate new input channel dimensions. Again, MPGD exhibits advantages over other candidate solutions by a large margin.

*4.2.5 Convergence Progress* We visualize the training progress of each task in Figure 2, where MSE, Shrinkage and MPGD are displayed for comparison. As can be seen, the training curve of MPGD overlaps with MSE at early training stage, when the model behaves poorly and almost all the entries in each data instance will be selected into $\mathcal{E}$ given a constant $\lambda$. Along with the training progress, the benefits of MPGD manifest and a lower loss can be observed. This is especially obvious on lithography and timing prediction tasks.

## 5 Conclusion

In this paper, we discussed the necessity of high-accurate predictive AI solutions in back-end chip design flows and the drawbacks of the well-adopted MSE cost function used in multiple prediction tasks. To further improve the reliability and generalization capability of these predictive AI models, we proposed a new training algorithm termed MPGD which incorporates an adaptive cost function and in-data active sampling. MPGD is able to achieve provably better convergence than traditional gradient descent with MSE loss. Served as a Plug-and-Play component, MPGD can in general boost the existing predictive AI performance in various physical design tasks.